\documentclass[twoside,11pt]{article}

\usepackage[final]{melba}  
\usepackage{float}

\hypersetup{
    colorlinks=true,
    citecolor=black,
    linkcolor=blue,
    urlcolor=black,
}

\newcommand{\figWidth}{0.80}
\usepackage{amsmath,amsfonts}

\usepackage{environ} 
\NewEnviron{HUGE}{%
    \scalebox{1}{$\BODY$} 
} 

\ShortHeadings{Impact of rater style on deep learning uncertainty}{Vincent et al.}
\firstpageno{1}

\title{Impact of individual rater style on deep learning uncertainty in medical imaging segmentation}

\author{\name Olivier Vincent$^{1,2}$ \email ovincent.poly@gmail.com \\  
	\AND
	\name Charley Gros$^{1,2}$ \email charley.gros@gmail.com \\
	\AND
	\name Julien Cohen-Adad$^{1,2,3}$ \email jcohen@polymtl.ca \\
\addr $^{1}$ NeuroPoly Lab, Institute of Biomedical Engineering, Polytechnique Montreal, Montreal, Canada\\ 
	\addr $^{2}$ Mila - Quebec AI Institute, Montreal, QC, Canada\\
\addr $^{3}$ Functional Neuroimaging Unit, CRIUGM, Université de Montréal, Montreal, Canada\\

}

\begin{document}

\maketitle

\begin{abstract}
While multiple studies have explored the relation between inter-rater variability and deep learning model uncertainty in medical segmentation tasks, little is known about the impact of individual rater style. This study quantifies rater style in the form of bias and consistency and explores their impacts when used to train deep learning models. Two multi-rater public datasets were used, consisting of brain multiple sclerosis lesion and spinal cord grey matter segmentation. On both datasets, results show a correlation ($R^2 = 0.60$ and $0.93$) between rater bias and deep learning uncertainty. The impact of label fusion between raters’ annotations on this relationship is also explored, and we show that multi-center consensuses are more effective than single-center consensuses to reduce uncertainty, since rater style is mostly center-specific.
\end{abstract}

\begin{keywords}
  Deep learning, Segmentation, Uncertainty, Rater style
\end{keywords}

\section{Introduction}

\subsection{Context}
Inter-rater variability limits the achievable segmentation performance of deep learning segmentation by introducing human error to the ground-truth \citep{carass_longitudinal_2017}. Tasks such as multiple sclerosis (MS) lesions segmentation are highly challenging due to the smallness of lesions and the poorly defined borders, leading to a low inter-rater agreement and high deep learning model uncertainty \citep{gros_automatic_2019,nair_exploring_2020}. For instance, some experts tend to over-segment, others under-segment, yielding “confusion” for the segmentation model trained on data labelled by different raters \citep{jungo_effect_2018}. Understanding the rater style could allow for better performance of models, e.g., by integrating this knowledge within the deep learning training scheme.


\section{Related Works}
Previous studies have shown that the rater style can be learned \citep{shwartzman_impact_2019}, and therefore the inter-rater disagreement patterns could potentially also be learned by the model \citep{zhou_exploring_2019}. There as also been work on jointly learning individual rater characteristic at the same time as "true" consensus segmentation, in classification \citep{zhang_disentangling_2020}, segmentation \citep{tanno_learning_2019} , and object detection \citep{sudre_lets_2019}. Also shown was that the method used to generate the ground truth from multiple rater annotations, e.g., label fusion \citep{jungo_effect_2018} and label sampling \citep{shen_improving_2019} largely impacts the model uncertainty.

\subsection{Contribution}
While many studies have addressed the uncertainty introduced by multiple raters, fewer work addressed the uncertainty introduced by a single rater. A model trained with data from a single rater will still exhibit some level of uncertainty due to rater style, and our goal is therefore to find what factors in a rater’s style generate uncertainty. Those factors could include tendency to under/over-segment, consistency across images, non-independence of raters (e.g. influence of the expert who trained the rater). Intuitively, a non-biased and highly consistent rater would be the ideal candidate for training a deep learning model. We therefore expect a correlation between a rater’s bias/consistency and the uncertainty of the model trained with their annotations. This would mean that characterization of the rater’s bias could eventually be incorporated as prior knowledge within the learning scheme or in the reporting of uncertainty (post-processing).

\section{Material and Methods}

\subsection{Datasets}
Two public MRI datasets with multiple raters annotations were used. The first is a brain multiple sclerosis (MS) lesion dataset introduced at a MICCAI 2016 challenge \citep{commowick_objective_2018}. It consists of 15 subjects each annotated by seven raters from three different centers. The second dataset is a spinal cord (SC) gray matter (GM) introduced at a segmentation challenge challenge \citep{prados_spinal_2017}, which includes 40 subjects with annotations from four raters (all raters from a different center).

\subsection{Metrics}
In this paper, we characterize rater's style using rater bias and consistency. Since the consensus of all raters is the closest we have to the real ground truth, we define a rater’s bias to be the average difference (in terms of positive voxel count) between the rater’s annotation and the consensus across all volumes:
\begin{align} \label{eq:bias}
    \text{bias} = \frac{\sum_{\text{images}} n_{\text{rater}}-n_{\text{consensus}}}{\text{\# of images}}
\end{align}
With $n_{\text{XX}}$ the number of positive voxels in a segmentation mask (i.e., belong to the target segmentation class). Consensus is defined by majority voting, as explained in section \ref{sec:processing}. A positive or a negative bias therefore measures if a given rater has a tendency to over- or under-segment, respectively. Images refer to 3D volumes, but using a 2D slices as a basis instead would give the same results up to a constant factor, since bias is an average.

Similarly, we define rater consistency as the standard deviation of the difference (in terms of positive voxel count) between the rater’s annotation and the consensus across all volumes:

\begin{align} \label{eq:consistency}
    \text{consistency} = \sqrt{\frac{\sum_{\text{images}} (n_{\text{rater}} - n_{\text{consensus}} - \text{bias})^2}{\text{\# of images}}}
\end{align}

Consistency therefore measures whether a rater is always either over-segmenting or under-segmenting  (consistent, close to zero) or if they are doing a bit of both (inconsistent: higher values).

 We choose to use an absolute bias metric as opposed to a relative one since we think not all slices deserved the same weight. For example, it would be unfair to penalize a rater by the same amount for a 10\% error on a slice showing only a single 10-voxel lesion, versus for a 10\% error on a slice with multiple lesions totaling hundreds of voxels. The error in the former case is likely negligible, whereas the error in the latter case is large and systematic (multiple lesions), but both would have the same impact on the computed bias if we had used a relative metric. We however considered using relative instead of absolute metrics, by normalizing the difference used in bias and consistency by the number of positive voxels in the consensus in each image. Results of these investigations are in appendix \ref{sec:relBias}, and show that the bias/uncertainty relationship in figure \ref{fig2} and \ref{fig3} still holds when using relative bias.

\subsection{Processing} \label{sec:processing}
Images were resampled ($1 \times1 \times 1mm^3$ for MS brain and $0.25 \times 0.25 \times 2mm^3$ for SC GM) and cropped (respectively $160 \times 224$ and $128 \times 128$) before being fed to the models. Data augmentation (rotation, translation, scaling) was applied slice-wise. Datasets were split $60/10/30$ randomly for training/validation/testing respectively. 2D U-Nets \citep{ronneberger_u-net:_2015} were trained slice-wise with the annotation of each individual rater. While it is no more state of the art, a 2D U-Net is sufficient since it can achieve near inter-rater variability levels of performance \citep{gros_automatic_2019,vincent_automatic_2020}. Additional performance would not be beneficial since the main goal is to study uncertainty and not segmentation performance. A more advanced architecture would probably only result in overfitting on some rater styles.  Training was done on NVIDIA P100 GPUs using the open source framework \href{http://ivadomed.org/}{\color{blue}ivadomed} \footnote{\url{http://ivadomed.org/}}  $v2.1.0$ \citep{gros_ivadomed_2021} which is based on PyTorch \citep{paszke_pytorch:_2019}. Configuration files containing all hyperparameters for both datasets are also available \href{https://github.com/olix86/paper_rater_uncertainty}{\color{blue}here} \footnote{\url{https://github.com/olix86/paper_rater_uncertainty}}. Models were trained using a Dice loss \citep{milletari_v-net_2016}. Inference was then done on the test set to measure model’s performance (Dice score) and aleatoric uncertainty \citep{wang_aleatoric_2019}. Uncertainty is estimated using test-time data augmentation (rotation, translation, scaling), and is computed as the entropy of 10 Monte Carlo samples for each image. The exact settings for the transforms are described in the config files linked above. The choice of aleatoric uncertainty was made because  it is considered as being representative of “inherent” uncertainty in the data, whereas epistemic uncertainty is considered dependent on the model parameters (i.e. it could go away with more data) \citep{kendall_what_2017,kiureghian_aleatory_2009}. All the previous steps (pre-processing, data augmentation, training, evaluation and uncertainty computation) were done with ivadomed. Preliminary experiments on the MS brain dataset showed that generating ground truth with STAPLE \citep{warfield_simultaneous_2004} yielded similar results in terms of rater style (bias \& consistency) compared to majority voting. The only difference was a constant offset to all raters bias, meaning that majority voting has a tendency to over-segment when compared to STAPLE. Since this affects all raters and doesn’t have an impact when comparing styles between raters, ground-truths were generated using majority voting due to it being easier to interpret (i.e. consensus voxel = 1 if at least 50\% of raters voted 1). By default, the term “consensus” will refer to this combination of all raters for a given dataset, however, single-center consensuses were also computed using the same method and will be compared to the global consensus.

\section{Results}
\subsection{Rater style}

\begin{figure}[h]
  \centering
      \includegraphics[width=\figWidth\textwidth]{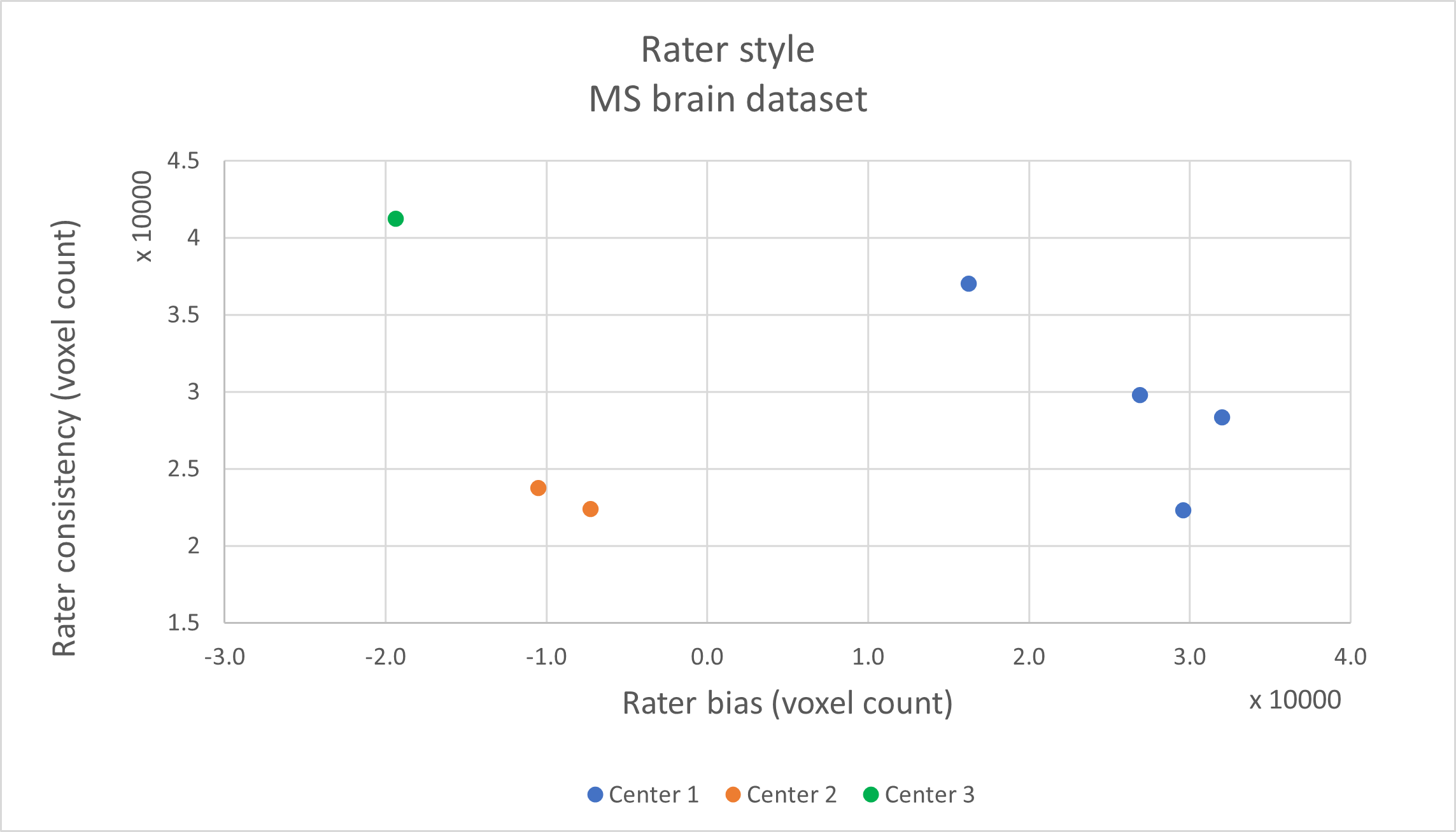}
  \caption{Rater style is characterised by consistency and bias. Dots colour corresponds to each rater’s center.}
  \label{fig1}
\end{figure}

We first examine rater style in the form of bias and consistency relative to consensus for MS brain. Styles are shown in Figure \ref{fig1} and solely depend on the ground truths from each rater; they do not involve any deep learning model.

We notice 3 clusters which are clearly delimited by the center to which raters belong. This implies that rater style depends a lot more on the rater’s center than its individual characteristics. Indeed, cluster radii of $[0, 1.8, 12] \times 10^3$ are a lot smaller than the distances between pairs of clusters centroid $[21, 36, 47] \times 10^3$. To assess the quality of the clustering we use the Davies-Bouldin index \citep{davies_cluster_1979}, a metric which quantifies the quality of clustering through ratios of intra-cluster scatter to inter-cluster distance (lower is better). Here, $DB=0.21$, meaning that intra-cluster scatter is quite lower than inter-cluster distances. Our hypothesis is that uncertainty for individual raters should follow a similar center-centric pattern assuming it depends on the rater style. This does not apply for the GM dataset since it contains only a single rater per center.

\subsection{Uncertainty}
\begin{figure}[h]
  \centering
      \includegraphics[width=\figWidth\textwidth]{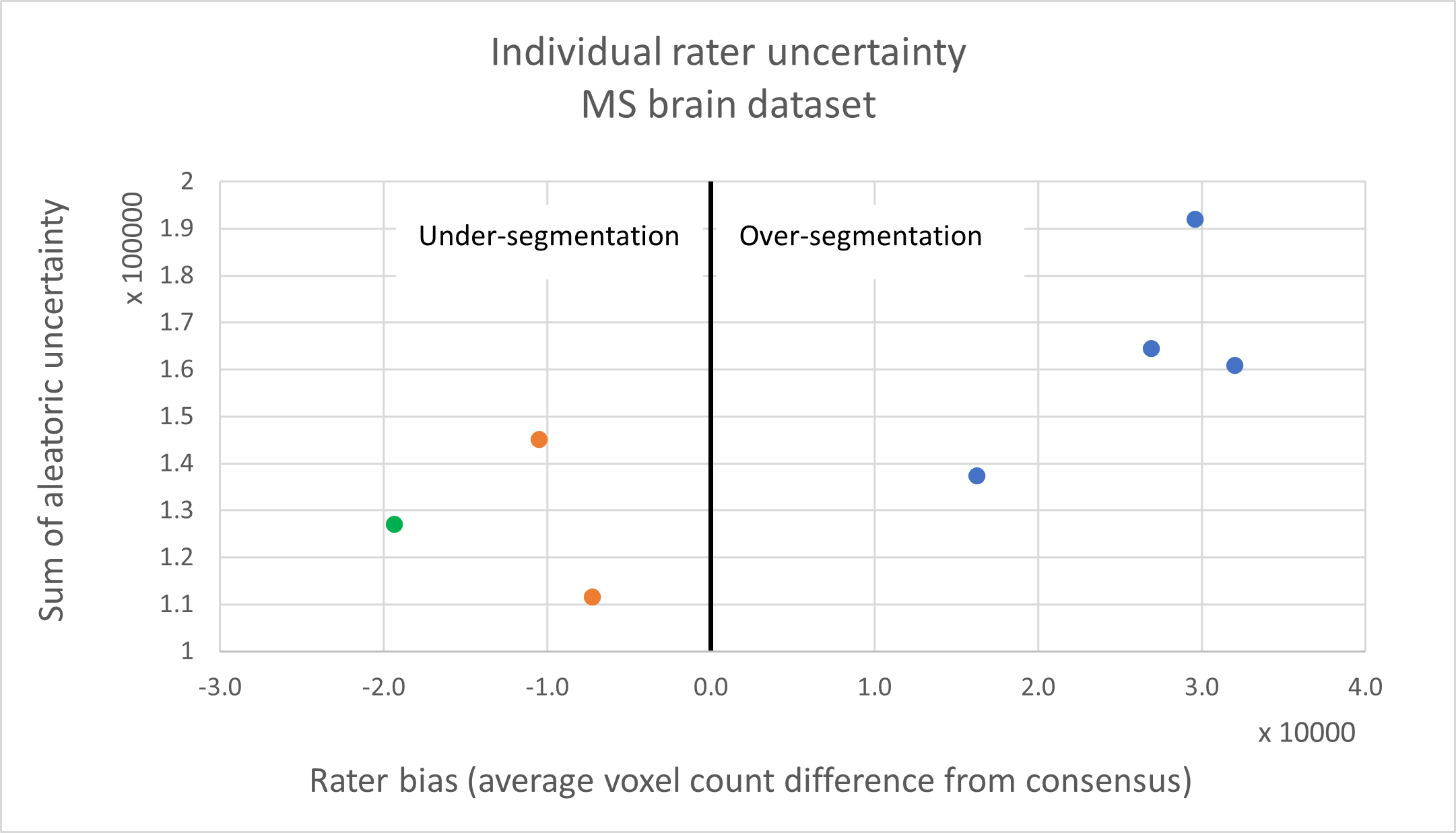}
  \caption{Relationship between the uncertainty of models trained for each rater and the bias of the corresponding rater for the MS brain dataset. Each colour corresponds to a center. $R^2 = 0.60$}
  \label{fig2}
\end{figure}

\begin{figure}[h]
  \centering
      \includegraphics[width=\figWidth\textwidth]{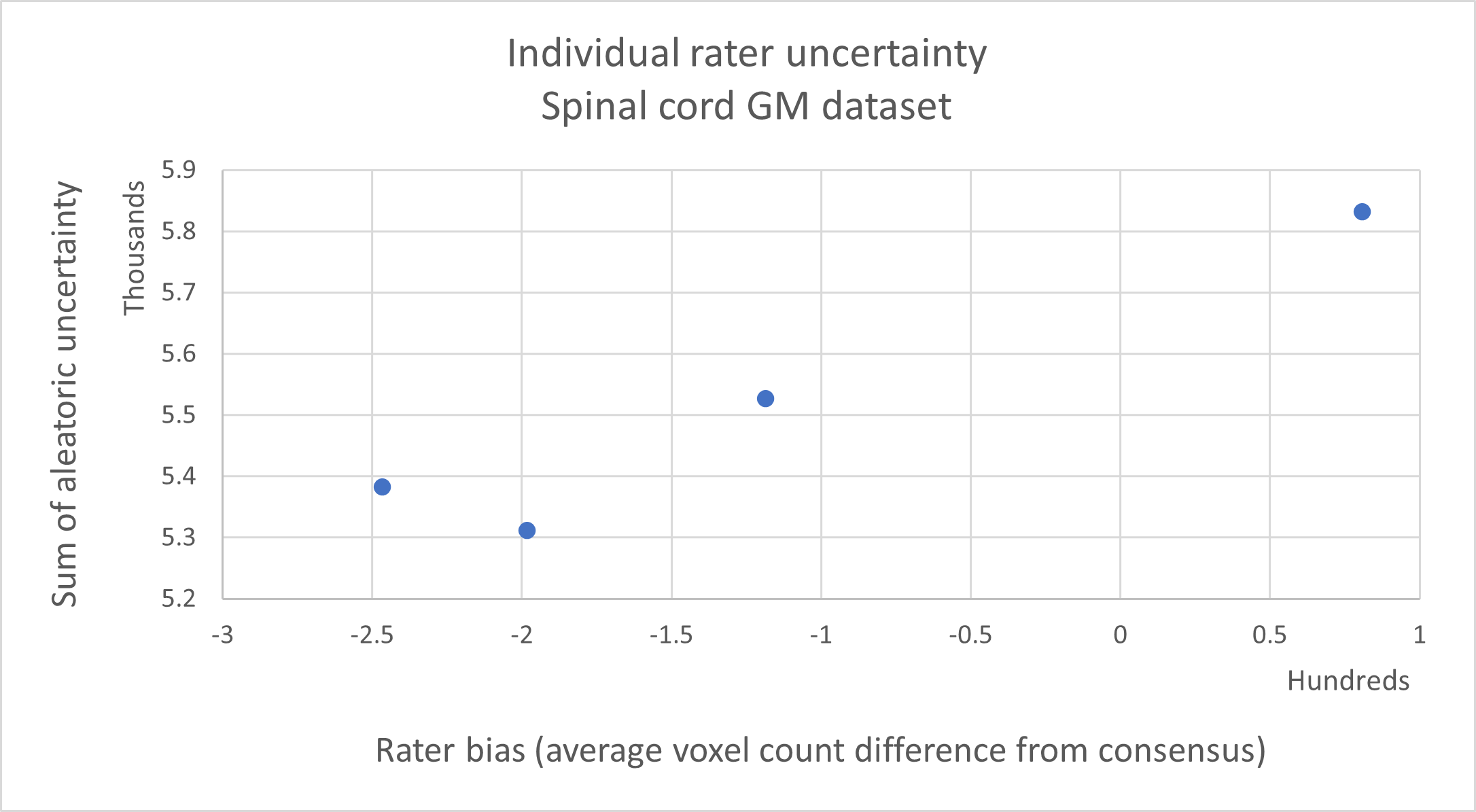}
  \caption{Relationship between the uncertainty of models trained for each rater and the bias of the corresponding rater for the SC GM dataset. $R^2 = 0.93$}
  \label{fig3}
\end{figure}

Next, we look at how uncertainty in models trained separately for each rater relates to rater style for both datasets, in Figure \ref{fig2} and \ref{fig3}.

In both datasets raters with a higher bias also have higher uncertainty. Over-segmentation (bias $>0$) seems to be associated with higher uncertainty than under-segmentation (bias $<0$). Raters are also clustered by center for the MS brain dataset, but in this case the distance between clusters is smaller than in the rater-style graph, since other factors also influence uncertainty, such as noise in data and the limited size of the training set.

\begin{figure}[h]
  \centering
      \includegraphics[width=\figWidth\textwidth]{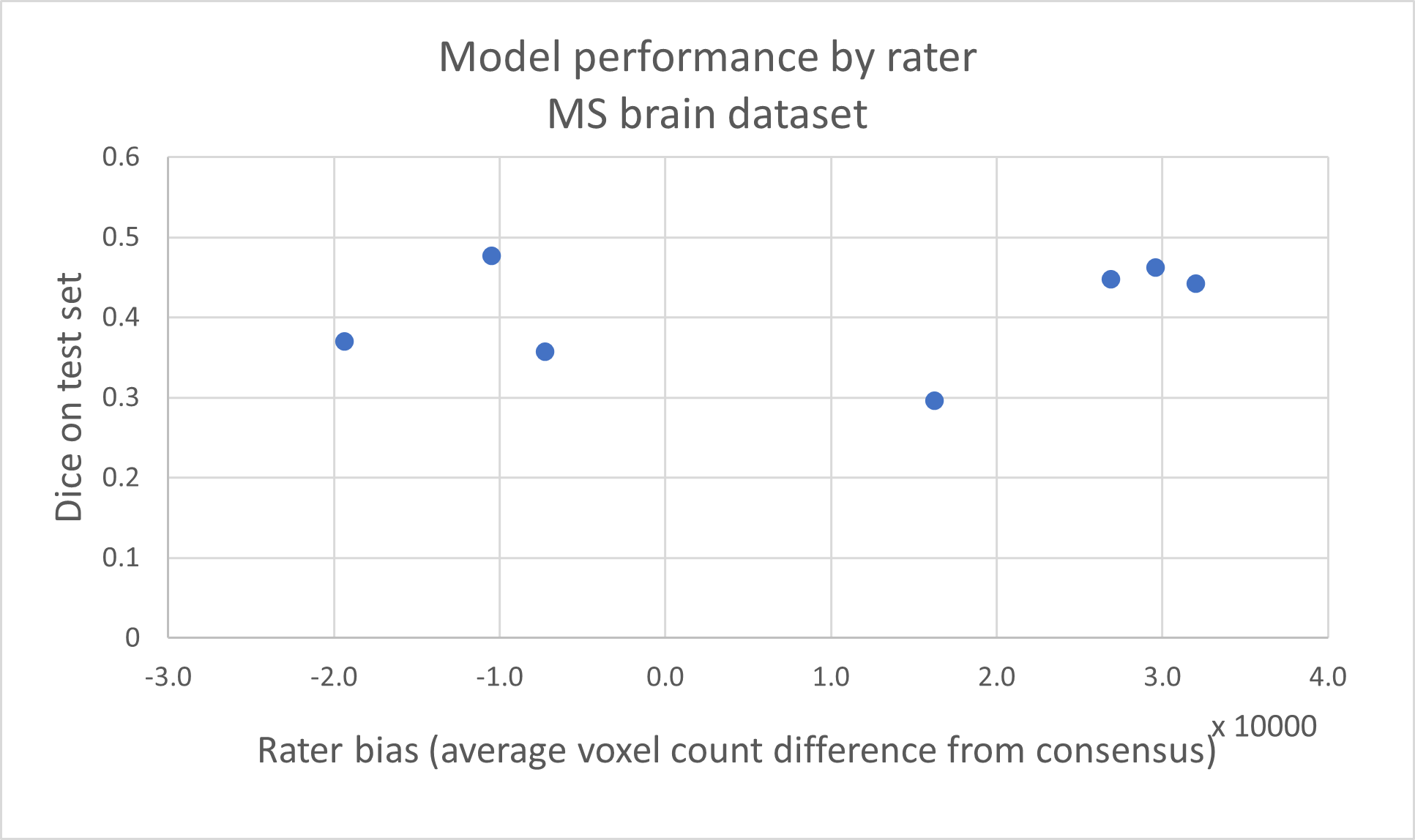}
  \caption{Relationship between the Dice score on the test set of models trained for each rater and the bias of the corresponding rater ($R^2 = 0.07$). Dice score is computed for each model on the same ground-truth that was used for training (e.g. model trained on data from rater $X$ is evaluated with respect to ground truth unseen during training from the same rater).}
  \label{fig4}
\end{figure}

It is interesting to note that while a higher rater bias produces higher uncertainty, it does not affect model performance as assessed by the Dice score  ($R^2 = 0.07$), as shown in Figure \ref{fig4}.

\subsection{Consensus}
\begin{figure}[h]
  \centering
      \includegraphics[width=\figWidth\textwidth]{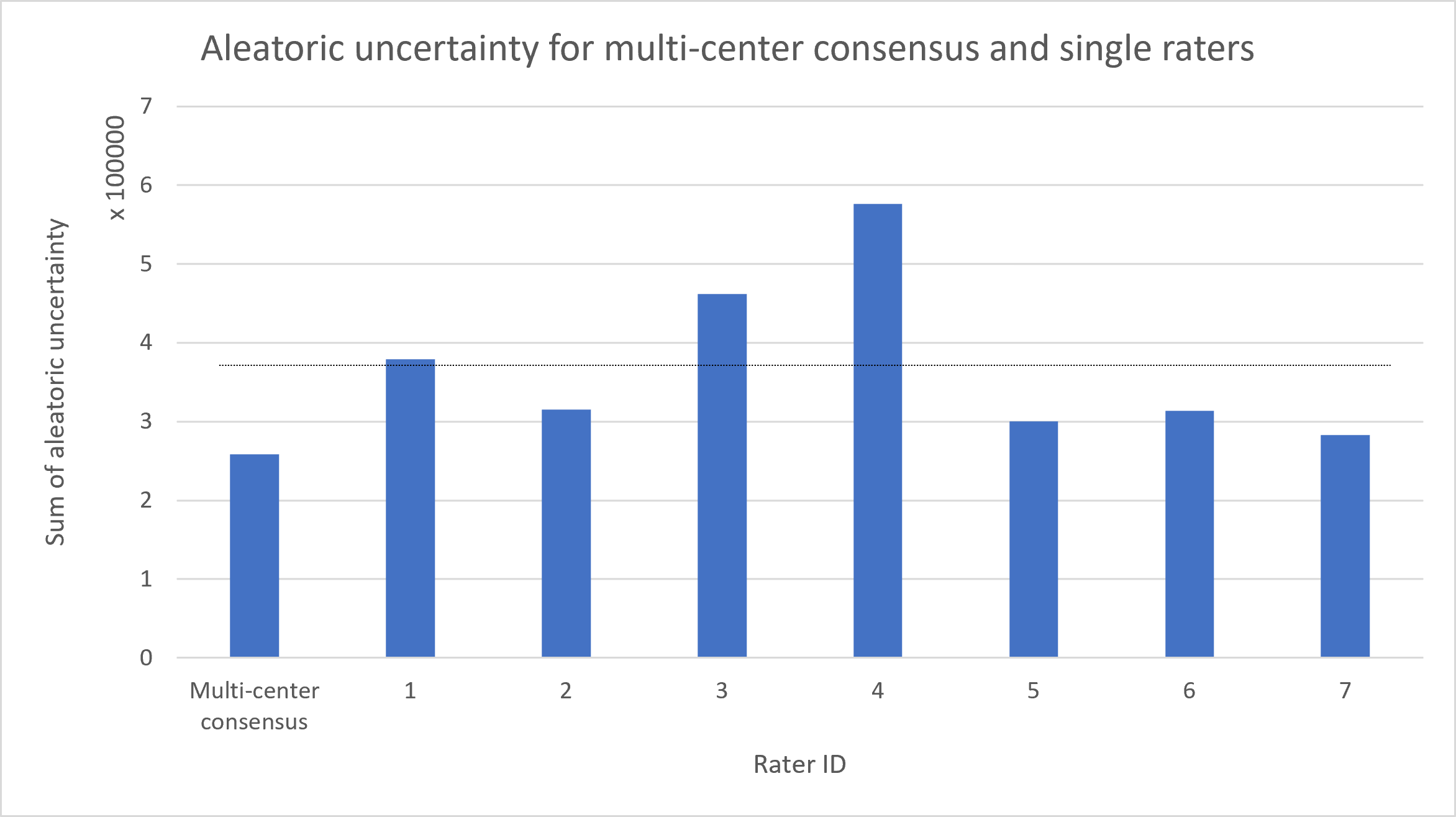}
  \caption{Comparison of uncertainty for individual raters and multi-center consensus for the MS brain dataset. Dotted line is the average of single raters uncertainty.}
  \label{fig5}
\end{figure}

All raters exhibit some level of bias and as we saw earlier, bias is correlated with uncertainty (Figures \ref{fig2}-\ref{fig3}). We now investigate whether combining raters through consensus would lower uncertainty when compared to single-rater training. Results of this investigation are shown in Figure \ref{fig5}, highlighting a consensus uncertainty 30\% lower than the average across individual raters. 

\begin{figure}[h]
  \centering
      \includegraphics[width=\figWidth\textwidth]{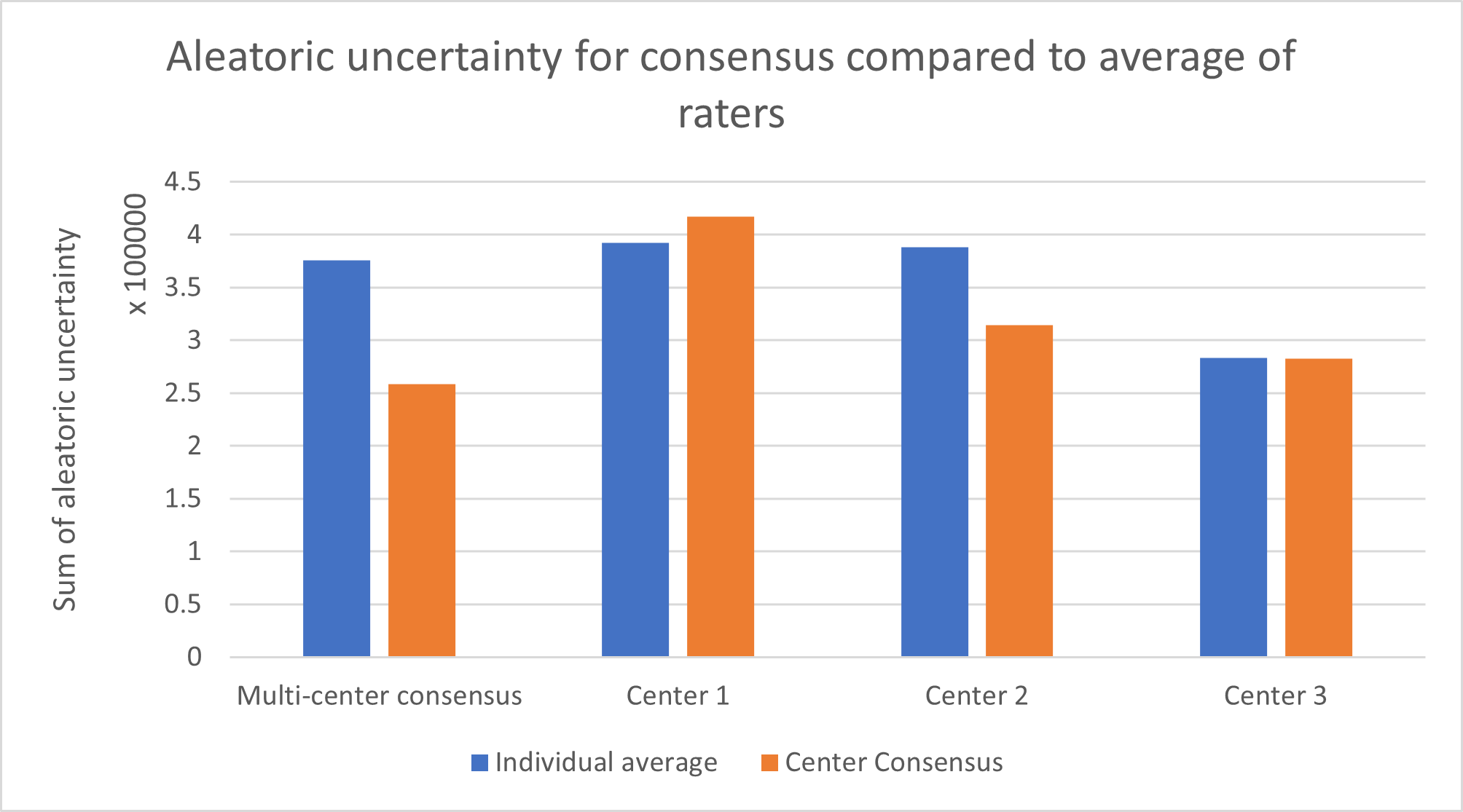}
  \caption{Per-center comparison of average rater uncertainty and consensuses uncertainty for the MS brain dataset. }
  \label{fig7}
\end{figure}

Center-wise consensuses do not, however, exhibit the same characteristic, as they have higher uncertainty than the global (multi-center) consensus and are comparable (lower for center 2, slightly higher for center 1, and irrelevant for the single rater of center 3) to the average uncertainty of their raters used individually (Figure \ref{fig7}).

Finally, the previous results are also reflected in the performance (Dice score) of models with the global consensus scoring a full $0.1$ above the average of individual raters, and scoring $[0.05 - 0.09]$ above single center consensuses as shown in Table \ref{tab1}. Thus, it seems that combining raters from different centers has a more positive impact on uncertainty and Dice than combining raters from the same center.

\begin{table}[h]
\centering
\begin{tabular}{|l|l|}
\hline
\multicolumn{2}{|c|}{Dice score for consensus} \\ \hline
Raters average                  & 0.42         \\ \hline
Center 1 consensus              & 0.47         \\ \hline
Center 2 consensus              & 0.46         \\ \hline
Center 3 consensus              & 0.43         \\ \hline
Multi-center consensus          & 0.52         \\ \hline
\end{tabular}
\caption{Dice score for different combinations of raters for the MS brain dataset. Dice score is computed for each model on the same ground-truth that was used for training (e.g. model trained on data from center $X$'s consensus is evaluated with respect to ground truth unseen during training from the same center).}
\label{tab1}
\end{table}

\section{Discussion}
\subsection{Key takeaways}

This study shows that rater style can be characterised by measuring rater consistency and bias. Moreover, results from the brain MS dataset suggest that rater style is mostly center-specific instead of rater-specific. Results also show on both MS brain and SC GM that when using annotations from a single rater to train a deep learning model, a high rater bias leads to high model uncertainty. This is interesting since these models are trained on annotations from a single rater and therefore have never “seen” the consensus although bias relative to consensus still impacts uncertainty. While rating style impacts the amount of uncertainty, bias doesn’t directly affect the average performance (Dice score) of the model meaning that the rater style can be learned by the model regardless of uncertainty. A mechanism that could potentially explain why oversegmentation leads to higher uncertainty is partial volume effect.  Indeed, a rater that undersegments (e.g. labelling only voxels that contain 100\% lesion tissue, and not those at the boundary that contain some other tissue) would give an easier task to the model; voxels labelled as lesions are homogeneous, and simple to identify. At the opposite, a rater that oversegments also includes voxels containing a varying percentage of lesion tissue, which is potentially harder since there is less homogeneity, therefore yielding more uncertainty.

Another interesting result is that uncertainty was lower for the global consensus model (i.e., when fusing all raters’ annotations into a single binary annotation) than for models trained using annotations from a single rater. We hypothesize this phenomenon originates from the biases of individual raters which get smoothed away when combining raters from different centers which have different styles. This is also probably why combining raters’ annotations from a single center (center-wise consensuses) does not reduce a model's uncertainty : individual bias can’t cancel out since we combine raters with similar styles and shortcomings. A single rater, such as the one from Center \#3, can therefore have lower uncertainty than the consensus from the four raters of Center \#1 due to their higher bias. Multi-center consensus could therefore be a mechanism to lower the impact of rater style.

This lower uncertainty for the global consensus however opens up questions regarding the impact of inter-rater variability on uncertainty. Inter-rater variability by definition is not present for single-rater models, but is present in center-wise consensuses, and is at its highest for the global consensus since it combines raters with diverging styles. Our results therefore suggest that the reduction in rater bias when going from single rater to global consensus has a bigger impact on uncertainty than the addition of inter-rater variability. It is therefore possible that inter-rater variability is indeed present but relatively constant throughout the dataset, thus not generating much uncertainty. A limitation of this study is that the number of raters (7 for the MS dataset and 4 for the SC GM dataset) is relatively small, therefore our results would benefit from further validations in datasets with larger pools of raters from different centers.

\vspace{-0.3cm}
\subsection{Impact and perspectives}

Raters used in the MS brain study were junior raters trained by senior raters from their center \citep{commowick_objective_2018}, therefore the mutual influence between raters during the learning and segmentation process probably drives the similarities in rating style. This center-wise rater style pattern raises a few questions concerning label fusion, which is largely used in deep learning medical imaging studies. Indeed, in the case of the MS brain dataset, since the split between centers is 4-2-1, if one uses a majority voting consensus it essentially becomes the vote of the four raters from a single center, negating the benefits of having two additional centers with raters in the study. It is doubtful that STAPLE and its variants could really solve the issue since it is based on majority voting, only with weights updated iteratively. If the four raters from one center dominate during the first iteration, the remaining raters will see their weighting be progressively reduced until convergence.

Future studies should therefore consider whether raters from the same center can really be considered independent, or if voting should be weighted by centers instead of raters. Weights of raters could also be considered as hyperparameters that can be optimised in order to minimize uncertainty. Alternatively, raters weight could be incorporated into the input ground truth segmentation using a “soft training” pipeline \citep{gros_softseg_2020}. While our rater style was defined as simple metrics independent of deep learning models, it would be interesting to see if learned rater style approaches \citep{zhang_disentangling_2020,tanno_learning_2019, sudre_lets_2019} show a similar relationship to uncertainty. 

Other potentially interesting metrics include measuring boundary difference instead of volume difference. An exemple would be the average symmetric surface distance (ASSD) which computes the average euclidean distance between the object boundaries across raters. This metric would be particularly relevant for the MS lesion task where there is a large heterogeneity of object shape, and therefore it could be interesting to complement the volume difference analysis with some shape analysis. Indeed, an increase of lesion radius (e.g. evenly adding 1 voxel along the lesion boundary) would have a different impact on the relative increase of the lesion volume if the lesion is small or large (e.g., 10-voxels vs. 100-voxels lesion). Therefore, from a "radius segmentation style" perspective, it could be said that our absolute metric over-weights large lesions at the expense of small ones, whereas it would be the opposite for our relative metric. However, measures based on boundaries also have drawbacks. It is possible that two raters segment the same lesion volume with a slightly different boundary (translation, change of shape, etc). Therefore such a metric would measure a bias even though there is none (there is indeed a disagreement, but not in the form of over/under-segmentation that we are looking for). To summarize, the main drawback of our volumetric bias is the possibility that it turns out to be non-linearly dependant on some other underlying bias (e.g. if it in facts depends on the radius). While we present two bias metrics here, a detailed comparisons with other relevant metrics would be interesting to explore in future research.

Finally, uncertainty could have potential applications for quality control such as identifying biased raters when there are not many ratings available. As an example, a rater generating significantly higher than expected uncertainty for a given task could be excluded as an outlier. Conversely, rating style could be used as a pre-processing step to “correct” biases on an individual rater basis or incorporated as a prior in future deep learning architecture. Model segmentation could be modulated using metrics about the rater style (e.g used as inputs for FiLM layers \citep{perez_film:_2017,lemay_benefits_2021}).


\newpage
\acks{The authors would like to thank Andréanne Lemay and Lucas Rouhier from the IVADO medical imaging team for helpful discussions. Funded by the Institut de valorisation des données (IVADO), the Canada Research Chair in Quantitative Magnetic Resonance Imaging [950-230815], the Canadian Institute of Health Research [CIHR FDN-143263], the Canada Foundation for Innovation [32454, 34824], the Fonds de Recherche du Québec - Santé [28826], the Natural Sciences and Engineering Research Council of Canada [RGPIN-2019-07244]. FRQNT Strategic Clusters Program (2020‐RS4‐265502 ‐ Centre UNIQUE  ‐Union Neurosciences \& Artificial Intelligence –Quebec, Canada First Research Excellence Fund through the TransMedTech Institute. C.G has a fellowship from IVADOMED [EX-2018-4], O.V. has a fellowship from NSERC, FRQNT and UNIQUE.}

%
\ethics{The work follows appropriate ethical standards in conducting research and writing the manuscript, following all applicable laws and regulations regarding treatment of animals or human subjects.}

\coi{We declare we don't have conflicts of interest.}


\newpage
\appendix 
\section*{Appendix A. Relative bias}
\label{sec:relBias}

In this appendix we present the equivalent of figure \ref{fig2} and \ref{fig3} using relative instead of absolute bias. Relative bias is defined in equation \ref{eq:rel_bias} in a similar way as as absolute bias  in equation \ref{eq:bias}, with the only change being the fact that we normalize the difference between rater and consensus by the number of positive voxels in the consensus in each image, therefore ensuring no volume has a disproportionate weight.

 \begin{align} \label{eq:rel_bias}
    {\text{relative bias} = \frac{\sum_{\text{images}} \cfrac{n_{\text{rater}}-n_{\text{consensus}}}{n_{\text{consensus}}} }{\text{\# of images}}}
\end{align}

\begin{figure}[h]
  \centering
      \includegraphics[width=\figWidth\textwidth]{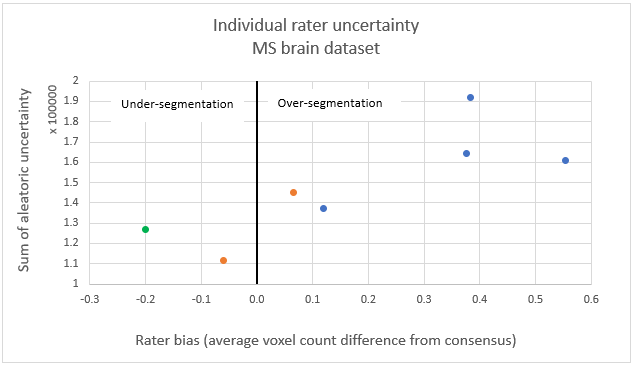}
  \caption{Relationship between the uncertainty of models trained for each rater and the relative bias of the corresponding rater for the MS brain dataset. Each colour corresponds to a center. $R^2 = 0.64$}
  \label{fig:unc_bias_ms_rel}
\end{figure}

\begin{figure}[h]
  \centering
      \includegraphics[width=\figWidth\textwidth]{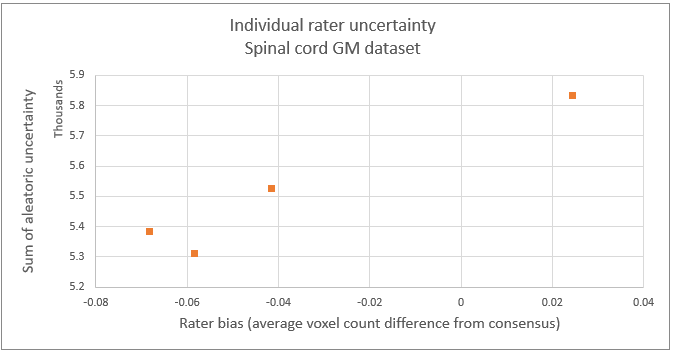}
  \caption{Relationship between the uncertainty of models trained for each rater and the relative bias of the corresponding rater for the SC GM dataset. $R^2 = 0.93$}
  \label{fig:unc_bias_gm_rel}
\end{figure}

Figure \ref{fig:unc_bias_ms_rel} and \ref{fig:unc_bias_gm_rel} show that on both datasets, the relationship between uncertainty and bias is still present using relative bias. Correlation is slightly stronger (0.64 vs 0.60) for MS and identical for (0.93) for GM than when using relative bias compared to absolute. On the qualitative side, for MS lesions we observe in figure \ref{fig:unc_bias_ms_rel} that one rater is an outlier (blue dot close to orange ones). It already was relatively far from its peer in figure \ref{fig2}, but this is exacerbated here. On GM segmentation, figure \ref{fig:unc_bias_gm_rel} shows that switching from relative to absolute bias makes pretty much no difference. The rater distribution is almost identical to figure \ref{fig3}. Overall GM bias is a lot lower than MS in both the absolute and relative cases, since the task is easier there is less disagreement between raters.  The lack of difference between the relative and absolute bias for GM is potentially explained by the fact that the GM volume varies a lot less across slices and subjects than MS lesions. It is inline with our expectations that relative bias is useful to accentuate the errors on very small lesions, which is why this re-weighting affects mostly the MS dataset.

\newpage
\vskip 0.2in
\bibliography{sample}

\end{document}